\documentclass{article}
\usepackage{amsmath}
\usepackage{amssymb}
\usepackage{bm}
\usepackage{authblk}
\usepackage{fancyhdr}
\usepackage{hyperref}
\newcommand{\citep}{\cite}
\DeclareMathOperator{\E}{\mbox{E}}
\DeclareMathOperator{\diag}{diag}
\renewcommand\vec{\bm}
\begin{document}

\title{Generalized Principal Component Analysis}
\author{F. William Townes\\
        Department of Biostatistics\\
        Harvard University\\
        \texttt{ftownes@g.harvard.edu}
        }

\lhead{F.W. Townes 2019}
\rhead{GLM-PCA}

\maketitle
\tableofcontents

\section{Introduction}

Principal component analysis (PCA) \citep{hotelling_analysis_1933} is widely used to reduce the dimensionality of large datasets. However, it implicitly optimizes an objective function that is equivalent to a Gaussian likelihood. Hence, for data such as nonnegative, discrete counts that do not follow the normal distribution, PCA may be inappropriate. A motivating example of count data comes from single cell gene expression profiling (scRNA-Seq) where each observation represents a cell and genes are features. Such data are often highly sparse ($>90\%$ zeros) and exhibit skewed distributions poorly matched by Gaussian noise. To remedy this, Collins \citep{collins_generalization_2002} proposed generalizing PCA to the exponential family in a manner analogous to the generalization of linear regression to generalized linear models. Here, we provide a detailed derivation of generalized PCA (GLM-PCA) with a focus on optimization using Fisher scoring. We also expand on Collins' model by incorporating covariates, and propose post hoc transformations to enhance interpretability of latent factors.

\section{Generalized linear models}

Generalized linear models (GLMs) are widely used for regression modeling when the outcome variable does not follow a normal distribution. For example, if the data are counts, a Poisson or negative binomial likelihood  can be used. Let $Y$ be the outcome variable. A fundamental aspect of GLMs is that the noise model is assumed to follow an exponential family likelihood:
\[\log f_Y(y;\theta) = c(y) + y\theta - \kappa(\theta)\]
In this formulation, $\theta$ is called the natural parameter and $\kappa(\theta)$ is called the cumulant function. The natural parameter is implicitly a function of the mean $\theta = \theta(\mu)$. The derivatives of the cumulant function yield moments. The mean is $\kappa'(\theta) = \mu$ and the variance is $\kappa''(\theta)$. Let $\rho(\mu)$ represent the variance function. It can be shown that the derivative of the natural parameter with respect to the mean is the inverse of the variance function:
\[\frac{d\theta}{d\mu} = \frac{1}{\rho(\mu)}\]

In regression modeling, the mean is an invertible, nonlinear function of the covariates and coefficients. The inverse of this function is called the link function: $g(\mu) = x'\beta$. Therefore, the GLM framework for regression involves maximizing the likelihood of the data $(y_i,x_i)$ with respect to the unknown vector of regression coefficients $\beta$. The most widely used algorithm for this optimization is a second-order method called Fisher scoring. For more details on GLMs, refer to \citep{agresti_foundations_2015}.

\section{GLM-PCA}

Suppose we have no covariates ($x$ is unknown) and $y$ is multivariate. Let $y_{ij}$ indicate the outcome of observation $i$ and feature $j$, with $i=1,\ldots,N$ and $j=1,\ldots,J$. In scRNA-Seq $i$ indexes over cells and $j$ indexes over genes. The GLM-PCA model, like PCA, seeks to reduce the dimensionality of the data $y_{ij}$ by representing it with an inner product of real-valued factors $u_i\in\mathbb{R}^L$ and loadings $v_j\in\mathbb{R}^L$. The number of latent dimensions is specified in advance as $L$. Let $r_{ij}=u_i'v_j$ be the real-valued linear predictor, $\mu_{ij} = g^{-1}(r_{ij})$ the mean, and $\theta_{ij} = \theta(\mu_{ij})$ the natural parameter. We define the derivative of the inverse link function as
\[h_{ij} = h(r_{ij}) = \frac{d\mu_{ij}}{dr_{ij}} = \frac{dg^{-1}(r_{ij})}{dr_{ij}}\]
The likelihood of the data is
\[\mathcal{L} = \sum_{i,j} c(y_{ij}) + y_{ij}\theta_{ij}-\kappa(\theta_{ij})\]
For numerical stability, we use a penalized likelihood as the objective function to be maximized:
\[\mathcal{Q} = \mathcal{L} -\frac{1}{2}\sum_{i,l}\lambda_{ul}u_{il}^2 -\frac{1}{2}\sum_{j,l}\lambda_{vl}v_{jl}^2\]
where $\lambda_{ul}$ and $\lambda_{vl}$ are small, non-negative penalty terms for $l=1,\ldots,L$.
The gradient is given by
\[\frac{d\mathcal{Q}}{d u_{il}} = \sum_j \frac{y_{ij}-\mu_{ij}}{\rho(\mu_{ij})}h_{ij}v_{jl} - \lambda_{ul}u_{il}\]
Applying the chain rule, the Fisher information is given by
\begin{align*}
-\E\left[\frac{d^2\mathcal{Q}}{du_{il}^2}\right] &= -\sum_j\E\left[\frac{\rho(\mu_{ij})(-1)-(y_{ij}-\mu_{ij})\psi'(\mu_{ij})}{\big(\rho(\mu_{ij})\big)^2}\big(h_{ij}^2v_{jl}^2\big)+\frac{y_{ij}-\mu_{ij}}{\rho(\mu_{ij})}\left(\frac{d^2\mu_{ij}}{dr_{ij}^2}v_{jl}^2\right)\right] + \lambda_{ul}\\
&= \sum_j\frac{h_{ij}^2 v_{jl}^2}{\rho(\mu_{ij})} + \lambda_{ul}
\end{align*}
Let $w_{ij} = 1/\rho(\mu_{ij})$. The Fisher scoring update for $u_{il}$ is given by
\[u_{il}\gets u_{il} + \frac{\sum_j (y_{ij}-\mu_{ij})w_{ij}h_{ij}v_{jl} - \lambda_{ul}u_{il}}{\sum_j w_{ij}h_{ij}^2v_{jl}^2 + \lambda_{ul}}\]
By a symmetric argument, the update for $v_{jk}$ is given by
\[v_{jl}\gets v_{jl} + \frac{\sum_i (y_{ij}-\mu_{ij})w_{ij}h_{ij}u_{il} - \lambda_{vl}v_{jl}}{\sum_i w_{ij}h_{ij}^2u_{il}^2 + \lambda_{vl}}\]
Since this update rule does not take into account any of the mixed second partial derivatives such as $d^2\mathcal{Q}/du_{il}dv_{jl}$ in computing the Fisher information, it is technically not true Fisher scoring but rather a diagonal approximation. This is actually an advantage since the true Hessian's dimension would be too large to efficiently invert. Note that blockwise coordinate ascent is also possible by vectorizing the updates across rows and/or columns, for example, let $u^{(l)} = (u_{1l},\ldots,u_{Nl})$ and $v^{(l)} = (v_{1l},\ldots,v_{Jl})$. Let $Y$ be the $J\times N$ data matrix with features as rows and observations as columns such that $y_{ij}$ is in column $i$, row $j$. Let $M$, $W$, and $H$ be similarly defined $J\times N$ matrices.
\begin{align*}
u^{(l)}&\gets u^{(l)} + \frac{\big((Y-M)\odot W\odot H\big)' v^{(l)} - \lambda_{ul}u^{(l)}}{\big(W\odot H^2\big)'\big((v^{(l)})^2\big)+\lambda_{ul}}\\
v^{(l)}&\gets v^{(l)} + \frac{\big((Y-M)\odot W\odot H\big) u^{(l)} - \lambda_{vl}v^{(l)}}{\big(W\odot H^2\big)\big((u^{(l)})^2\big)+\lambda_{vl}}
\end{align*}
where $\odot$ indicates elementwise multiplication, division is elementwise, and $H^2=H\odot H$.

This is a generic formulation. In special cases the update equations simplify considerably. For example, consider the canonical link function $g(\mu_{ij})=\theta(\mu_{ij})$ which implies $h_{ij}=\rho(\mu_{ij})=1/w_{ij}$. In this case the gradient becomes
\[\frac{d\mathcal{Q}}{du_{il}} = \sum_j (y_{ij}-\mu_{ij})v_{jl}-\lambda_{ul}u_{il}\]
and the Fisher information becomes $\sum_j \rho(\mu_{ij})v_{jl}^2 + \lambda_{ul}$.

\section{GLM-PCA with covariates}

So far we have implicitly assumed that for all dimensions $l=1,\ldots,L$, both $u^{(l)}$ and $v^{(l)}$ are unknown parameters to be estimated. In practice, row (feature-level) and/or column (observation-level) covariates may also be available. For example, in scRNA-Seq column covariates could indicate batch membership or cell cycle indicators which we want to regress out. Row covariates could include spline basis functions modeling gene-specific GC bias. Even if no covariates are available, simply incorporating a vector of all ones as a column covariate induces a row-specific intercept term, which is analogous to centering by feature in PCA.

Let $\tilde{U}\in\mathbb{R}^{N\times L}$ be the matrix whose columns are $u^{(l)}$ and $\tilde{V}\in\mathbb{R}^{J\times L}$ be the matrix whose columns are $v^{(l)}$. As before let $Y$ be the $J\times N$ data matrix. Suppose we are provided observation (column) covariates as a design matrix $X\in\mathbb{R}^{N\times K_o}$ and feature (row) covariates as a design matrix $Z\in\mathbb{R}^{J\times K_f}$. In addition, we consider the offset vector $\vec{\delta}\in\mathbb{R}^N$ (if no offset is needed, set $\vec{\delta}=\vec{0}$). We define the $J\times N$ real-valued linear predictor matrix as
\[R = AX'+Z\Gamma'+\tilde{V}\tilde{U}'+\vec{1}\vec{\delta}'\]
where $A\in\mathbb{R}^{J\times K_o}$ and $\Gamma\in\mathbb{R}^{N\times K_f}$ are matrices of regression coefficients and $\vec{1}$ is a vector of length $J$ with all ones. Now define the augmented column and row matrices as $U=\big[X,\Gamma,\tilde{U}\big]\in\mathbb{R}^{N\times(K_o+K_f+L)}$ and $V=\big[A,Z,\tilde{V}\big]\in\mathbb{R}^{J\times(K_o+K_f+L)}$ such that $R=VU'+\vec{1}\vec{\delta}'$. We define the following sets of dimensionality indices: $\Omega_o=\{1,\ldots,K_o\}$, $\Omega_f=\{K_o+1,\ldots,K_o+K_f\}$, $\Omega_L=\{K_o+K_f+1,\ldots,K_o+K_f+L\}$ and $\Omega=\Omega_o\cup\Omega_f\cup\Omega_L$. The set of column indices in $U$ that can be updated is $\Omega_u=\Omega_f\cup\Omega_L$, and for $V$ the updateable index set is $\Omega_v=\Omega_o\cup\Omega_L$.

To update $U$, for all $k\in\Omega_u$ do:
\begin{align*}
R &\gets VU'+\vec{1}\vec{\delta}'\\
M &\gets g^{-1}(R)\\
W &\gets \frac{1}{\rho(M)}\\
H &\gets h(R)\\
U_{[:,k]}&\gets U_{[:,k]}+ \frac{\big((Y-M)\odot W\odot H\big)' V_{[:,k]} - \lambda_{uk}U_{[:,k]}}{\big(W\odot H^2\big)'\big(V_{[:,k]}^2\big)+\lambda_{uk}}
\end{align*}
In general it is not necessary to penalize the regression coefficients, so if $k\in\Omega_f$, we may set $\lambda_{uk}=0$. To update $V$, for all $k\in\Omega_v$ do:
\begin{align*}
R &\gets VU'+\vec{1}\vec{\delta}'\\
M &\gets g^{-1}(R)\\
W &\gets \frac{1}{\rho(M)}\\
H &\gets h(R)\\
V_{[:,k]}&\gets V_{[:,k]}+ \frac{\big((Y-M)\odot W\odot H\big) U_{[:,k]} - \lambda_{vk}V_{[:,k]}}{\big(W\odot H^2\big)\big(U_{[:,k]}^2\big)+\lambda_{vk}}
\end{align*}
Where $\lambda_{vk}$ may be set to zero whenever $k\in\Omega_o$. At this point, all unknown parameters have been updated, so the objective function $\mathcal{Q}$ can be evaluated and monitored for convergence.

As previously stated, the above procedure is a diagonal approximation to full Fisher scoring. Alternating between full Fisher scoring of $U$ and $V$ is likely to be computationally unstable, since there is feedback between updating the unknown latent factors $\tilde{U}$ and the unknown loadings $\tilde{V}$. However, full Fisher scoring as a subroutine can be used to update $A = V_{[:,\Omega_o]}$ and $\Gamma=U_{[:,\Omega_f]}$, since there is no feedback in updating the corresponding fixed covariate matrices $X=U_{[:,\Omega_o]}$ and $Z=V_{[:,\Omega_f]}$. For example, to update $A$, for each $j=1,\ldots,J$ do
\[A_{[j,:]}'\gets A_{[j,:]}' + \big(X'\diag\left\{W_{[j,:]}\odot H_{[j,:]}^2\right\}X\big)^{-1}X'\diag\left\{W_{[j,:]}\odot H_{[j,:]}\right\}(Y_{[j,:]}-M_{[j,:]})\]
This can be used to show that ordinary GLM regression is a special case of GLM-PCA with covariates (namely, the case where $J=1$, $Z=\vec{0}$, and either $\tilde{U}=\vec{0}$ or $\tilde{V}=\vec{0}$). However, due to the inversion of a $K_o\times K_o$ matrix separately for all $J$ features, it is computationally demanding.

As an illustrative example of using covariates, consider a matrix of count data $Y$ with features in rows and observations in columns where the total counts in each column are not of interest (that is, the counts are only interpretable on a relative scale). We recommend setting the offset $\vec{\delta}$ to some constant multiple of the column sums of $Y$ such as the column means. Also recommended is to include feature-specific intercept terms by setting $X=\vec{1}$. The intercept terms will then be given by the (single column) matrix $A$. The number of latent dimensions $L$ should be chosen by the same methods used to determine the number of principal components in PCA.

\section{Rotation of latent factors to orthogonality}

Once the GLM-PCA objective function has been optimized on a dataset, postprocessing can improve interpretability of the latent factors. The first step, which we call the projection step, removes all correlation between latent factors and covariates without changing the predicted mean values $M=g^{-1}(R)$. Let $P_x=X(X'X)^{-1}X'$ and $P_z=Z(Z'Z)^{-1}Z'$ be projection matrices. Then the following reparametrization leaves $R$, and hence $M$ invariant (we omit the offset $\vec{\delta}$ for clarity):
\begin{align*}
R &= A X'+Z\Gamma'+\tilde{V}\tilde{U}'\\
&= Z\Gamma'+A X'+\tilde{V}\tilde{U}'X(X'X)^{-1}X'+\tilde{V}\tilde{U}'(\mathbb{I}-P_x)\\
&= Z\Gamma'+\big(A+\tilde{V}\tilde{U}'X(X'X)^{-1}\big)X'+\tilde{V}\tilde{U}'(\mathbb{I}-P_x)\\
&= \big(A+\tilde{V}\tilde{U}'X(X'X)^{-1}\big)X' + Z\Gamma' + Z(Z'Z)^{-1}Z'\tilde{V}\tilde{U}'(\mathbb{I}-P_x)+(\mathbb{I}-P_z)\tilde{V}\tilde{U}'(\mathbb{I}-P_x)\\
&= \big(A+\tilde{V}\tilde{U}'X(X'X)^{-1}\big)X' + Z\big(\Gamma + (\mathbb{I}-P_x)\tilde{U}\tilde{V}'Z(Z'Z)^{-1}\big)'+(\mathbb{I}-P_z)\tilde{V}\tilde{U}'(\mathbb{I}-P_x)
\end{align*}
Based on this, the first step in postprocessing is to set
\begin{align*}
A&\gets A+\tilde{V}\tilde{U}'X(X'X)^{-1}\\
\Gamma&\gets \Gamma + (\mathbb{I}-P_x)\tilde{U}\tilde{V}'Z(Z'Z)^{-1}\\
\tilde{U}&\gets (\mathbb{I}-P_x)\tilde{U}\\
\tilde{V}&\gets (\mathbb{I}-P_z)\tilde{V}
\end{align*}
As an example, consider the case where $X=\vec{1}$, so $A$ is a vector of feature-specific intercept terms. Then $P_x \tilde{U}$ computes the column means of $\tilde{U}$ and $(\mathbb{I}-P_x)\tilde{U}$ is a matrix whose column means are all zero. In this way, including feature-specific intercepts is analogous to centering the data prior to applying PCA. Both methods produce latent factors whose means are zero.

The second step in postprocessing, which we call the rotation step, is to rotate the factors so that the loadings matrix will have orthonormal columns. Let $\tilde{V}'=FD\hat{V}'$ be a singular value decomposition (SVD). By definition, $\hat{V}$ has orthonormal columns and we set this as the updated loadings matrix. Since $\tilde{V}\tilde{U}'=\hat{V}\big(DF'\tilde{U}'\big)$, we set $\hat{U}=\tilde{U}FD$ as the updated latent factors matrix. Note that if $\tilde{U}$ has column means of zero, then so does $\hat{U}$. PCA also produces an orthonormal loadings matrix.

The final postprocessing step is to rearrange the latent dimensions in decreasing magnitude, just like PCA orders principal components in decreasing variance. The L2 norm of a vector $x\in\mathbb{R}^n$ is defined as $\Vert x \Vert_2 = \sqrt{\sum_{i=1}^n x_i^2}$. Whenever the empirical mean of $x$ is zero, its empirical standard deviation equals its L2 norm divided by the constant $\sqrt{n-1}$. Therefore, ordering dimensions by L2 norm is equivalent to ordering by variance as long as the column means are zero. As a result of the previous step, all columns of $\hat{V}$ have L2 norm of one, so the magnitude of each dimension can be computed solely from the columns of $\hat{U}$. For each $l=1,\ldots,L$, compute $\Vert \hat{U}_{[:,l]}\Vert_2$. Then, arrange the columns of both $\hat{V}$ and $\hat{U}$ in decreasing order according to these L2 norms.

The postprocessing steps are computationally efficient so long as the numbers of latent dimensions $L$ and covariates $K_o,K_f$ are not too large. Specifically, the step is $\mathcal{O}\big(\max\{L,K_o,K_f\}^3\big)$ due to the matrix inversions and does not actually instantiate any large dense matrices like $R$ or $M$. Since our proposed Fisher scoring optimizer does not involve momentum terms that span iterations, it would be possible to perform the projection and/or rotation steps prior to convergence of the algorithm. For example, they could be run after every tenth iteration. However, this would reduce computational speed. Also, the postprocessing steps have no effect on predicted mean values $M$, and hence do not improve the theoretical goodness of fit to the data. The only benefit would be if the reduced correlation between dimensions improved numerical stability.

\bibliography{refs}

\end{document}